# Prevent Car Accidents by Using AI

Sri Siddhartha Reddy
*George Mason University*
sgudemup@gmu.edu

Yen Ling Chao
*George Mason University*
ychao4@masonlive.gmu.edu

Lakshmi Praneetha Kotikalapudi
*George Mason University*
lkotikal@gmu.edu

Ebrim Ceesay
*George Mason University*
eceesay@gmu.edu

*Abstract*—Transportation facilities are becoming more developed as society develops, and people's travel demand is increasing, but so are the traffic safety issues that arise as a result. And car accidents are a major issue all over the world. The cost of traffic fatalities and driver injuries has a significant impact on society. The use of machine learning techniques in the field of traffic accidents is becoming increasingly popular. Machine learning classifiers are used instead of traditional data mining techniques to produce better results and accuracy. As a result, this project conducts research on existing work related to accident prediction using machine learning. We will use crash data and weather data to train machine learning models to predict crash severity and reduce crashes.

## I. INTRODUCTION

The Association for Safe International Road Travel showed around 37000 persons die in automobile accidents each year, with another 2.35 million wounded or incapacitated by car accidents. Children under the age of 15 were responsible for 1600 deaths, while approximately 8000 persons between the ages of 16 and 20 were killed in automobile accidents [1]. As a result, vehicle accidents have become a societal hazard and one of the four leading causes of mortality in the metropolitan population. Every year, traffic accidents cost citizens hundreds of billions of dollars. Most of the losses were caused by a few big incidents. The purpose of major accident prevention is to avoid potentially hazardous road conditions. We can take appropriate actions and better manage financial and human resources if we can identify the primary reasons for catastrophic accidents. This case study's data is a nationwide accident data collection encompassing 49 states in the United States. From 2016 to 2020, there will be 1.5 million road accidents.

### A. Problem Statement:

The goal of this paper is to conduct a statistical analysis of the data, investigate the states with the most accidents, investigate when accidents are most likely to occur and the weather conditions at the time of accidents, and create a visual display: summarize and analyze the information, tell the overall situation of accidents in the United States, and discover the factors influencing the occurrence and severity of accidents; finally, the severity of the accident is predicted and evaluated.

### B. Significance:

This research studies the current state of road accidents in the United States and offers recommendations for reducing the number and severity of accidents based on several data parameters. We shall examine this data from many perspectives. First, examine the distribution of accidents in the United States based on geography. Second, from a time perspective, examine the period of concentrated accidents and the volatility of total accidents in recent years. Next, the weather dimension is used to examine the diverse effects of various meteorological elements on severity. Finally, analyze the POI variables that may be easily increased and decreased to raise and decrease the accident rate from the POI dimension, and give appropriate adjustment suggestions. As a result, in this study, we will utilize Python's panda module to evaluate and comprehend if current road accidents in the United States are rising or decreasing year by year, as well as which period is the peak period during which traffic accidents are most likely to occur. A simultaneous accident is a severe type of accident. Most of the accidents happen in either good or bad weather. We can clarify the causes of road accidents and minimize the accident rate based on the findings of these data analyses.

### C. The Reason for using Big Data Solution:

Road safety has become a key focus of current social concerns as the number of traffic accidents has increased. The location of the accident, the time of day, the driver's emotions, the weather, and other unpredictable and complicated elements all play a role in road accidents [2].

## II. LITERATURE REVIEW

Researchers used meta-analysis to assess the impact of bad weather on accident rates, demonstrating that weather has an impact on traffic safety. Extreme weather in the United States, according to the study, will have an impact on injuries and automotive accidents, notably on wet and snowy days, with the average percentage of collision rate on rainy and snowy days being 71 percent and 84 percent, respectively [3]. The severity of traffic accidents was assessed and forecasted using K-means clustering. It is typical to apply different machine learning methods in other research areas to find a solution for a problem and compare the results. This study compared the performance of two machine learning applications, random forest (RF) and Bayesian additive region trees (Bart), to investigate alternative machine learning approaches for predicting the severity of traffic accidents. It is discovered that, when compared to the prediction of the Bayesian additive region trees (Bart) model, the random forest (RF) model with meteorological circumstances has a greater prediction probability in estimating the severity of a traffic collision. The variable import ance technique reveals that the mode of traffic accidents and weather circumstances are two major factors that might give significant information in the modeling and estimating processes [4].

Their study analyzes data from 50 states to discuss traffic accident patterns and causes, as well as what can be done to prevent them, for US government agencies and the general public. Several variables that related to the severity of the accident were examined and analyzed using logistic regression [5].

Their analysis includes the number of accidents by year, number of accidents by state, the best time to travel by month, day, and hour, accident-prone areas in each state, factors that cause accidents such as weather, wind flow, temperature, location, and so on, deaths in each state, age groups of fatalities, drivers involved in accidents, drivers age groups, vehicles involved in accidents, and drivers who have consumed alcohol. Tableau was used to create the analysis platform. [6]

Another study tries to address this issue and go deeper into the elements that contribute to the rise in the number of car accidents. The data used in this study was collected constantly in the United States from 2016 to 2020 from traffic accident incidents captured by the Department of Transportation, law enforcement agencies, and traffic cameras. Two models were used to anticipate the impact of car accidents on traffic, with an emphasis on the primary causes of traffic accidents. The primary two factors determining car accident rates, according to the findings, are traffic induced by work rush hour and population density. [7]

Another research study was to present a model that can be used to explain why different countries have different rates of road deaths. As potential variables, national infrastructure, transportation, and socioeconomic indicators from international databases were explored. Stepwise regression analyses were used to create the model. [8]

R language demonstrates how data is related by analyzing traffic statistics and graphs. Locations of the accidents and the accident thermal chart were obtained after data preprocessing and data selection using R language Remap package remap and remap functions. In addition, to model the data, they used decision trees, linear regression, and the random forest approach. We can check the model's correctness and obtain the most accurate model based on the actual findings, which will help in predicting the model's accuracy with comparable data in the future. After validating the model and examining the data properties and relationship between the variables, the ultimate purpose of data analysis is to identify the most correct model and machine learning applications can prove this fact and NLP models could help to improve the model [9]–[18] [19]–[25] [26]–[30].

## III. Aims and Objectives

Traffic accidents are the greatest cause of death worldwide, taking the lives of millions of people each year due to their regularity. As a result, technology that predicts traffic accidents or accident-prone areas may be able to save lives. Nowadays, there is an increasing emphasis on traffic accident data mining and analysis, which can improve in-depth investigation and reduce traffic-related deaths.

We will use Python pandas to analyze car accidents in the United States in this project. Real-time data, accident sites, casualty analysis, driving speed, traffic conditions, road structure, and weather may all be used to anticipate accidents. Therefore, we can predict accidents based on a variety of factors. We'll consider factors such as road conditions, speed limits, and the state in which the accident occurred. The analysis of historical accident data will assist in determining the likely causal relationship between these factors and road accidents, enabling the creation of accident predictors to reduce the risk of injury caused by accidents. As a consequence, utilizing this data collection, a machine learning model is built and applied that can accurately anticipate when and where accidents will occur, reducing the number of automobile accidents.

## IV. Methodology

The vehicle accident dataset was obtained from the Sobhan Moosavi website, which has 1.5 million entries. To undertake data analysis, the dataset must be preprocessed and has to be cleaned up by eliminating null values and filling in blanks, which will aid in data normalization. Data insights may be obtained for analysis and better decision-making using a variety of machine learning methods. Charts, graphs, and ordered tables can be used to visualize the data. One of the most significant jobs in road accident analysis is to predict the severity level of the event using various classifier methods such as logistic regression, decision trees, and random forest.

## V. Proposed Approach To Solve Problem

### A. Focus:

Accidents in the United States may be utilized for several purposes, including real-time accident prediction, investigating the location of accident hotspots, fatality analysis and deriving causal principles to anticipate accidents, and researching the potential influence of climate on accidents. The analysis's goal is as follows:

1) Which state has the highest number of accidents in the United States?
2) What time of day is the most prone to accidents, and what is the pattern of accident change?
3) What connection exists between the accident and the weather?

### B. Steps:

The project is broken into five stages:

*1) Problem Definition:* Every year, traffic accidents account for a high share of serious injuries reported. However, establishing the conditions that cause these occurrences is frequently difficult, making it more difficult for local law enforcement to handle the frequency and severity of traffic accidents. Many questions, however, remain unsolved. As a result, it is critical to forecast and investigate traffic accidents that we want to know which elements are the most important in causing automobile accidents and to understand which factors of the accident will have a bigger influence on traffic accidents.

*2) Data Collecting:* Data cleaning is an important phase in the data analysis process [19], and the quality of the findings is strongly related to the model impact and the conclusion. In practice, data cleansing often consumes 50% - 80% of the time allotted to the analysis process. Data cleaning can help to enhance data quality, reduce interference, and acquire useful and trustworthy data to aid decision-making.

*3) Data Transformation:* In order to structure the data for analytical usage, we built a database with Parquet format in an HDFS cluster as described in [**?**]. The performance results of PySPark APIs demonstrated by [**?**] show a significant speedup in the SQL API. We perform the data transformations that are required to partition and store the datasets using the PySpark API and make it available for model building.

*4) Data Cleaning:* Data cleaning is an important phase in the data analysis process, and the quality of the findings is strongly related to the model impact and the conclusion. In practice, data cleansing often consumes 50% - 80% of the time allotted to the analysis process. Data cleaning can help to enhance data quality, reduce interference, and acquire useful and trustworthy data to aid decision-making.

*5) Data Visualization:* Data visualization is a way of communicating information more simply and effectively by conveying facts through data, primarily using graphical approaches. The goal of visualization is to present facts more naturally, making the data more objective and compelling.

## VI. PRELIMINARY RESULTS

Exploratory data analysis has been done on the dataset where cleaning and preparation of the data and then analyze it with different plots and visualizations. In the data preparation, the dataset file has been loaded using pandas, and cleaning is done by fixing missing or incorrect values. Some of the columns that we analyzed were State, start time, Start Lat, Start Long, Temperature, and Weather condition to analyze the severity of the accidents state-wise weather-wise. Some of the results produced on analysis are below.

The top 10 weather conditions at the time of the accident were displayed in a bar chart. We can see that the severity of accidents is highest when the weather is clear, followed by cloudy weather. As a result, it's worth noting that the vast majority of accidents have a severity of 2, that the vast majority of accidents happen in "Clear" weather, and that the overall number of accidents surpasses 1,200,000.

From the map, we see that California is recorded to be the highest accident-prone state.

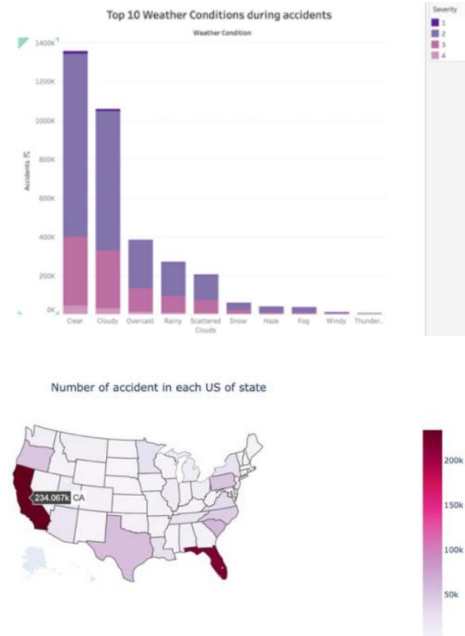

According to this graph, the top ten states have the highest rate of auto accidents. As can be seen, the top three states with the most accidents are California, Florida, and South Carolina, all of which are at the top of the US in terms of population and GDP. The expansion of traffic will be fueled by the economy's prosperity, and the likelihood of traffic accidents will rise as well.

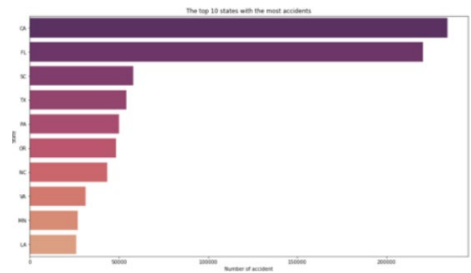

The Pie chart depicted the number of accidents that occurred between 2016 and 2021. According to the findings, the number of car accidents in 2021 would be much greater than the number in 2016-2020, which was 1,511,745. The second highest number of accidents will occur in 2020, with a total of 625,864. During COVID-19, Americans traveled less, the number of automobiles on the road reduced, and many people began to speed up. As a result, the abrupt spike in traffic accidents in 2021 resulted in an increase in traffic accidents.

The higher the number of accidents in the morning peak, the greater the traffic flow; and as the evening peak comes to an end, the serious accident rate continues to rise, reaching a peak at 17:00 and then gradually decreasing; it is speculated that this is due to fatigue driving after normal working hours. As a result, fatigue is more likely than low visibility to cause serious accidents.

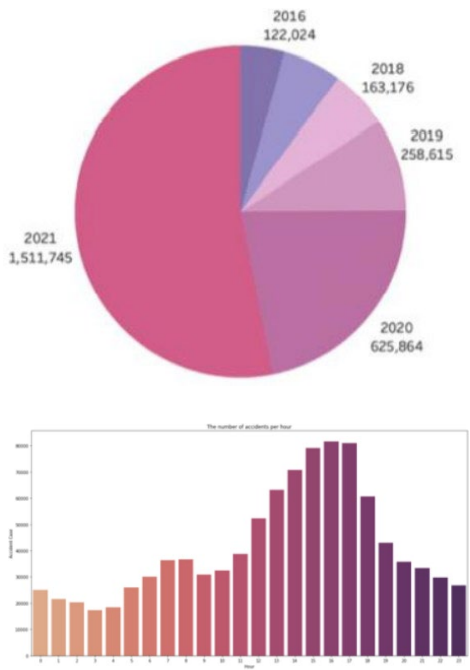

2020, resulting in 52,668 injuries and 847 fatalities. However, the number of collisions on Virginia's roadways fell in 2020, while road deaths increased to the highest level in more than a decade. According to the research findings, there will be a 2.4 percent rise in accident deaths in 2020 compared to 2019, while speed-related fatalities would climb by 16.3 percent compared to last year . According to the Virginia Highway Safety Office, the top three counties with the most fatalities in 2020 will be Fairfax County, Chesterfield County, and Henrico County.

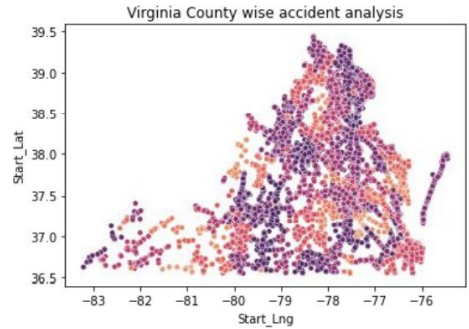

Considering the climate conditions such as temperature and pressure, on analyzing them, can infer that all the severity 2 is more when the pressure is between 29.5 and 30.5 and found that other severities are recorded to be more at that particular pressure point.

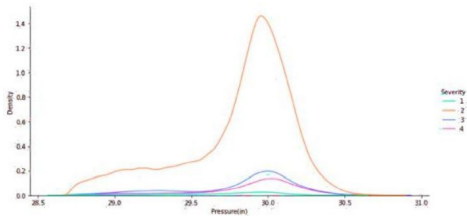

Considering temperature as a factor, severity 2 is highest when the temperature is recorded between 40F and 80F.

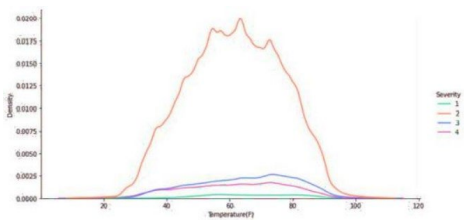

Furthermore, the repercussions of an automobile accident may be life-changing, including catastrophic injury, death, and a significant financial burden. A driver's degree of experience or age, attention, drunkenness, reactivity to weather conditions, or high-risk conduct such as speeding are all factors that lead to an accident. According to the Virginia Highway Safety Office, there were 128,172 events in 2019, with 65,708 injuries and 827 fatalities. However, there were 105,600 accidents in

## VII. MODEL ANALYSIS

By building a model using logistic regression, decision trees, and random forests, and compare the results to see which model can predict the severity of a vehicle accident better. The most often used evaluation indicators (Metric) in classification tasks are accuracy, precision, recall, and F score, with accuracy being the most generally utilized, followed by recall rate. The accuracy rate, which is the number of all prediction pairings divided by the total number, is the easiest to comprehend. Simultaneously, the accuracy rate measures the proportion of predicted positive samples in all projected positive samples, and the recall rate calculates the proportion of predicted positive samples in all genuine positive samples. As a result, the greater the recall rate, the better the model's capacity to locate positive samples, and precision and recall have a harmonized average.

| Model | Logistic Regression | KNN | Random Forest |
|---|---|---|---|
| Accuracy | 0.71 | 0.75 | 0.82 |
| Precision | 0.69 | 0.78 | 0.88 |
| Recall | 0.68 | 0.78 | 0.76 |
| F1-score | 0.66 | 0.79 | 0.79 |

The ROC curve is commonly used in machine learning to evaluate the classifier's reliability in binary classification situations. The Precision-Recall curve can reveal more information when working with data sets that are severely unbalanced. The classifier labels the categorization of an instance as positive or negative in binary classification problems, which can be represented by a confusion matrix. The x-axis of the ROC

|  | Actual Positive | Actual Negative |
|---|---|---|
| Predicted Positive | TP | FP |
| Predicted Negative | FN | PN |

curve is FPR, while the y-axis is TPR. The chance of being wrongly forecasted as a positive sample in the actual negative sample is denoted by FPR; TPR denotes the likelihood of being accurately predicted in the actual positive sample. The x-axis of the Precision-Recall curve represents recall, while the y-axis is accuracy and Precision-Recall curve are therefore used to assess the classification performance of a machine learning system on a particular data set. Each data set has a predetermined number of positive and negative samples. The ROC curve and the Precision-Recall curve have a strong connection.

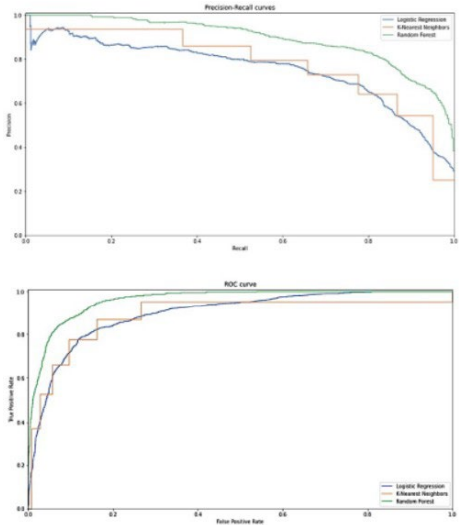

We discovered that the random forest has the best outcomes after analyzing and comparing the results of the three models using the accuracy, ROC, and PR curves. Through a random forest visualization of feature importance, we can see which traits are most likely to relate to severity, and which are least likely to be correlated. As a result, we may utilize the significance score to decide which attributes to eliminate when they are irrelevant or which ones to preserve when they are relative. This is a feature choice that can help to simplify the problem being modeled, speed up the modeling process, and, in certain situations, increase the model's performance.

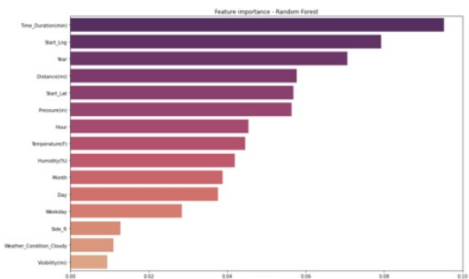

VIII. CONCLUSION

Traffic safety has always been a significant problem in the development of sustainable transportation and predicting the severity of traffic accidents remains a critical challenge in the field of traffic safety. As a result, a considerable fraction of the number of people wounded in traffic accidents is recorded year after year. However, pinpointing the particular factors that led to these occurrences is frequently difficult, making it more difficult for local governments to reduce the frequency and severity of traffic accidents. The current study was conducted to discover which elements, such as specific aspects of the weather conditional, have a significant influence in triggering automobile accidents. The second goal of this study was to look at the impacts of forecasting the severity of automobile crashes using machine learning and focus on in Virginia State, with the goal of minimizing the number of car accidents.

The severity of road accidents may be predicted using the three types of traffic accident severity prediction models developed in this work. Second, the data from traffic accidents includes many dimensions, and the random forest technique is well-suited to multi-dimensional data and can produce accurate predictions. We can determine if these factors are influence by predicting the severity of traffic incidents. Based on the result weather conditions, accident time and location, lighting conditions, and other factors all have an impact on the severity of traffic accidents. As a result, there is a useful measure for preventing or lessening the severity of road accidents. These elements can be studied and planned for as much as feasible while developing traffic roadways to lessen the influence on traffic accidents.

With significant road accident rates on the rise, there is a pressing need to address the issue, and big data is emerging as a potent tool for reducing the number of people killed or injured in car accidents. As the number of deaths continues to rise, so do the monetary losses caused by accidents. We propose that predictive analytics, such as machine learning and advanced big data systems, be used to collect key insights about car crashes, such as where, when, why, and how serious the crash was, data that is required to create predictive crash maps that analyze historical and recent data to identify high-risk areas. Drivers might be warned to be extra cautious in certain locations using predictive collision maps. Local governments can take the required actions to improve road safety at the same time.

Furthermore, enhancing our road infrastructure may have a huge influence on the safety of both drivers and pedestrians. We believe that programmed stop lights can improve traffic flow and reduce the risk of accidents. The number of fatalities might be dramatically decreased by utilizing smart software to guide traffic at congested crossings. The location, direction, speed, and accident history of each automobile may offer extensive information on traffic flow and circumstances, prevalent accident hotspots, and how drivers react to various scenarios. These findings allow government infrastructure officials to undertake large road upgrades, enhance overall driving conditions, and minimize human error.

## IX. Future Work

Predicting automobile accidents is a critical area for future study as a method of reducing accidents. Only using three modeling algorithms which are logistic regression, KNN, and random forest, are employed in this study to predict accidents severity in Virginia State. However, more algorithms can be investigated for verification in the future to improve prediction results. While we can look at the effect of states on the severity of car accidents in the future work. These data may be used to compute the relevant statistics of states as well as the severity or number of accidents to examine whether there is a link between the time of year and the severity of accidents. As a result, it is advised to conduct more research and pay closer attention to the severity of automotive accidents in each state. Alternatively, we may include the model into a real-time accident risk prediction model, allowing us to investigate the detailed link between some critical components and accident severity in greater depth. [31], [31], [32], [32], [33], [33], [34], [34], [35], [35], [36], [36], [37]